\pdfoutput=1
\documentclass[conference]{IEEEtran}
\IEEEoverridecommandlockouts

\usepackage{cite}
\usepackage{amsmath,amssymb,amsfonts}
\usepackage{algorithmic}
\usepackage{hyperref}
\usepackage{graphicx}
\usepackage{textcomp}
\usepackage{xcolor}
\def\BibTeX{{\rm B\kern-.05em{\sc i\kern-.025em b}\kern-.08em
    T\kern-.1667em\lower.7ex\hbox{E}\kern-.125emX}}
    
\begin{document}

\title{An AI-Powered Autonomous Underwater System for Sea Exploration and Scientific Research}

\author{
\IEEEauthorblockN{2\textsuperscript{nd} Mariam Al Nasseri}
\IEEEauthorblockA{\textit{College of Technological
Innovation} \\
\textit{Zayed University}\\
Abu Dhabi, United Arab Emirates \\
202111707@zu.ae.ac}
\and
\IEEEauthorblockN{1\textsuperscript{st} Hamad Almazrouei}
\IEEEauthorblockA{\textit{College of Technological
Innovation} \\
\textit{Zayed University}\\
Abu Dhabi, United Arab Emirates \\
201912368@zu.ae.ac}
\and
\IEEEauthorblockN{3\textsuperscript{rd} Maha Alzaabi}
\IEEEauthorblockA{\textit{College of Technological
Innovation} \\
\textit{Zayed University}\\
Abu Dhabi, United Arab Emirates \\
202104533@zu.ae.ac}
}

\maketitle

\begin{abstract}
Traditional sea exploration faces significant challenges due to extreme conditions, limited visibility, and high costs, resulting in vast unexplored ocean regions. This paper presents an innovative AI-powered Autonomous Underwater Vehicle (AUV) system designed to overcome these limitations by automating underwater object detection, analysis, and reporting. The system integrates YOLOv12 Nano for real-time object detection, a Convolutional Neural Network (CNN) (ResNet50) for feature extraction, Principal Component Analysis (PCA) for dimensionality reduction, and K-Means++ clustering for grouping marine objects based on visual characteristics. Furthermore, a Large Language Model (LLM) (GPT-4o Mini) is employed to generate structured reports and summaries of underwater findings, enhancing data interpretation. The system was trained and evaluated on a combined dataset of over 55,000 images from the DeepFish and OzFish datasets, capturing diverse Australian marine environments. Experimental results demonstrate the system's capability to detect marine objects with a mAP@0.5 of 0.512, a precision of 0.535, and a recall of 0.438. The integration of PCA effectively reduced feature dimensionality while preserving 98\% variance, facilitating K-Means clustering which successfully grouped detected objects based on visual similarities. The LLM integration proved effective in generating insightful summaries of detections and clusters, supported by location data. This integrated approach significantly reduces the risks associated with human diving, increases mission efficiency, and enhances the speed and depth of underwater data analysis, paving the way for more effective scientific research and discovery in challenging marine environments.
\end{abstract}


\section{Introduction}

\subsection{Background}The sea remains one of the most mysterious and least explored regions of our planet. To begin with, extreme conditions, including high pressure, poor visibility, and unpredictable underwater landscapes, pose significant challenges to traditional exploration methods, making them dangerous, costly, and limited in effectiveness. In fact, despite covering \textbf{over 70\% of Earth's surface}, only \textbf{about 5\% of the ocean} has been fully explored, underscoring the vast unknowns that persist due to these challenging conditions \cite{21_noaa_2023_how}. In addition, human divers face substantial risks and data collection is often \textbf{slow} and \textbf{incomplete}. For example, typical dive durations are limited to \textbf{less than an hour} (approximately 50 minutes) due to air supply constraints, and marine data collected manually can take weeks to analyze and validate, significantly slowing scientific progress \cite{22_cameron_2025_operation}. However, recent advancements in \textbf{Artificial Intelligence (AI)} and autonomous systems are creating new opportunities. As a result, by combining \textbf{Computer Vision}, \textbf{Machine Learning}, and \textbf{Automated Reporting}, sea exploration can now become \textbf{safer}, \textbf{faster}, and \textbf{more insightful}. Therefore, AI-powered AUVs equipped with real-time image processing and autonomous navigation are already demonstrating superior performance in identifying marine species and environmental anomalies compared to traditional methods \cite{23_jain_2024_ai}.

\subsection{Project Overview} 
This project presents an \textbf{AI-powered Autonomous Underwater Vehicle (AUV) System} designed to transform sea exploration. The system integrates \textit{YOLOv12} (You Only Look Once) for real-time object detection with a \textit{Large Language Model (LLM)} that generates structured reports and summaries of underwater findings, focusing on detected objects and clusters. It is capable of detecting and assigning marine objects to pre-defined clusters, including both known and unknown types, and automatically produces detailed reports to support scientific research, environmental monitoring, and future ocean discoveries. Furthermore, \textit{K-Means} clustering is applied to analyze patterns in marine biodiversity, enabling improved classification and ecosystem understanding.

\subsection{Scope}
By deploying this autonomous system, the project aims to reduce the risks associated with human diving, increase the efficiency and speed of underwater missions, and overcome environmental challenges such as low visibility and physical obstacles. Autonomous systems are increasingly replacing human divers in hazardous environments, improving safety while significantly increasing the volume and speed of data collection\cite{24_bernier_2024_underwater}. Energy optimization is a key focus, enabling longer missions, while aiming to preserve detection accuracy and mission reliability\cite{34_yang_2019_energy}. The use of AI-generated summaries further enhances data analysis and supports more informed scientific research. As the system continues to evolve, future developments will include unsupervised learning capabilities such as automatic anomaly detection. These advances will \textbf{drive} the continued progress of intelligent, autonomous sea exploration.

\section{Literature Review}
Our project contributes to the advancement of \textbf{Autonomous Underwater Vehicle (AUV)} technology by developing an AI-powered system that automates underwater object detection and reporting,  To put this work into perspective, this literature review will look at previous studies and applications in underwater detection to emphasizing effective approaches to further enhance our research. Then we will review recent advancements and finally summarize the current state of the field and identify certain gaps.

    \subsection{Similar Implementations}
    Numerous recent studies have investigated various methods for applying AI to automate underwater object identification and interpretation. \textbf{DeepFins} project is among the most advanced and relevant in the field of underwater AI detection, this paper presented a combination of \textit{YOLOv11} and a custom designed motion segmentation module\cite{7_jalal_2025_deepfins}. The reason for this mix is that \textit{YOLO} is a deep learning object detector required for fast and accurate object detection, while the motion segmentation module is used to identify moving objects between frames which is much needed for tracking fish and other constantly moving targets underwater. Even in harsh underwater conditions, the hybrid approach achieves high detection accuracy by capturing both how objects appear and how they move over time . As for the results of this paper, \textbf{DeepFins} accomplished great results in terms of detection by reducing false positives of fast-moving fish, these results consist of an F1 score of 90.0 percent specifically on the \textbf{OzFish} dataset\cite{7_jalal_2025_deepfins}, the reason they were able to achieve this high F1 score was due to their customized use of motion-based segmentation alongside \textbf{YOLOv11} this optimizing detection for fast-moving objects. While their usage of \textbf{DeepFins} had very high results, it mainly focuses on detection accuracy it does not include K-means clustering nor \textit{Large Language Models (LLMs)} which are key aspect of our system.

Another direction that is similarly related is the development of \textbf{MERLION}, which targets underwater visual monitoring and uses these visuals to give an informative summary sample catered to the user needs such as coral reef views or fish activity. Although MERLION does not use \textit{LLM}, it is able to make a connection between visual content to textual context. The purpose of MERLION is to help researchers reduce the amount of video that needs to be reviewed which in return saves time and effort\cite{8_a2020_merlion}. Our system adds on this concept by using a \textit{Large Language Model (LLM)} to produce clear summaries in addition to identifying and classifying underwater items, further advancing the objective of converting raw underwater imagery into insightful information. Similarly a model name \textbf{MarineInst} which is specifically on underwater data as a \textit{Vision-language Model}, its main purpose is to use language models to describe detections\cite{9_zheng_marineinst}, which aligns with our project's goal of generating \textbf{natural language summaries} after detection and grouping.

Each of these systems contributes significantly to the implementation of an underwater AI system. Our system fills a gap left by no single existing system by merging detection (YOLO), grouping (K-means), and summarization (LLM) into a single pipeline. Together, these modules work alongside feature extraction (CNN) and dimensionality reduction (PCA) to show important components of the broader solution that our system is creating.

    \subsection{Recent Advancements}
    It is crucial to dive into the recent advancements especially since artificial intelligence has significantly enhanced the efficiency and accuracy of underwater object detection and assessment. The following sections examine the advancements in technologies such as \textbf{YOLO}, \textbf{Convolutional Neural Networks (CNNs)}, unsupervised techniques like \textbf{K-means clustering}, and \textbf{Large Language Models (LLMs)s} showing their importance to the pipeline developed in this study. 
    
        \subsubsection{Progress in YOLO and Real-time Object Detection}
        \textbf{YOLO} has made numerous advancements throughout the years improving speed and accuracy making it the driving force behind real-time object detection \cite{15_alrabbani_2025_yolov12}. Initial releases such as YOLOv2 and YOLOv3 improved accuracy for a range of object sizes by using deeper networks and broad feature extraction. YOLOv4 to YOLOv6 added data augmentation, multiple feature scales, and CSPNet \cite{3_tian_2025_yolov12}. YOLOv7 through YOLOv9 enhanced model scalability, this is to have consistent detection performance in GPU resource-limited devices and advanced computing systems, YOLOv10 and YOLOv11 enhanced these capabilities even further, improving the model's capacity to recognize intricate visual patterns in a variety of settings \cite{15_alrabbani_2025_yolov12}. As for YOLv12, it focuses on integrating efficient \textit{attention mechanisms} which is to improve detection of challenging objects without sacrificing speed. These mechanisms help the model focus more on the important parts of the image, such as objects, and ignore the less relevant areas to improve object detection without sacrificing speed, YOLOv12 also achieves a  \textbf{mAP score of 40.6\%} on the \textbf{COCO} dataset, further proving its detection improvements. These features align directly with our system’s need for accurate, real-time detection in unpredictable underwater scenes\cite{3_tian_2025_yolov12}.
        
        \subsubsection{Developments in CNN Architectures}
        The main method of deep learning is \textbf{Convolutional Neural Networks (CNNs)}. This is because, feature extraction is automated with CNN which minimizes human interaction and improves its learning ability\cite{16_jia_2022_underwater}. CNN-based models have been the reason that a large number of underwater object detection systems are more accurate and efficient. For instance due to stronger feature extraction, in the research done by Jia et al.\cite{16_jia_2022_underwater} \textbf{DenseNet} with YOLOv3 was usec to detect small underwater creatures, likewise \textbf{ResNet50} uses residual learning to extract deep features and enhancing training stability \cite{17_he_2015_deep}; this was groundbreaking but slow. To counteract this, \textbf{EfficientDet} was introduced showing excellent outcomes in underwater contexts with \textbf{87.7\% accuracy} while maintaining high speed \cite{16_jia_2022_underwater}. These advancements show how CNN architectures meet the demand for accuracy and efficiency which are key requirements for our project.
        
        \subsubsection{Applications of K-means Clustering in Feature Analysis}
        \textbf{K-means clustering} is a popular unsupervised learning algorithm to group data points based on the similarity of features and has been used in several marine image analysis applications\cite{27_farshadfarahnakian_2023_a}. In underwater image segmentation, a study modified and improved K-means to separate foreground objects from the background more significantly, even when conditions are challenging with low contrast and blurred edges. Centroid initialization was tackled using a strategy comparable to \textbf{K-means++}, which in turn results in more stable segmentation for underwater scenes and more accurate separation of marine life from surrounding environments\cite{29_chen_2021_an}. Another study applied \textbf{U-Net} modeling results in feature extraction based on grain-size characteristics, followed by \textit{K-means} classification of these into five sediment classes such as gravel and silt, corresponding to well-known sediment maps derived from sonar data. This application highlights the ability of clustering to translate continuous data into discrete, meaningful marine categories\cite{25_zhou_2023_classification}. A further study on \textit{K-means} has also been used in the behavioral analysis context for detecting anomalous patterns of vessels by clustering ship trajectories in order to identify “dark” ships with the help of \textbf{Automatic Identification System (AIS)} data. The method was efficient in detecting subtle patterns in an unsupervised manner, offering insight into hidden behavioral trends in maritime activity\cite{27_farshadfarahnakian_2023_a}. \textbf{K-means++} is used in our system to cluster the outputs from \textit{YOLO} object detection, which allows recurring elements like types of fish, or objects to be clustered together; thus, allowing further analysis and helping the language model generate meaningful summaries by organizing similar objects into structured clusters for interpretation and reporting.
        
        \subsubsection{The Role of Large Language Models}
        A \textbf{Large Language Model (LLM)} which is a type of machine learning algorithm can generate text that appears human-written based on a large amount of text data that it is trained on \cite{18_aws_2023_what}. The use of LLMs in underwater detection has arisen to assist in turning raw visual data into text resembling human writing. A study was conducted using GPT-4V tested marine tasks such as identifying fish and calculating reef coverage, but findings indicate that their domain-specific accuracy is often lacking \cite{19_zheng_2024_exploring}. To counteract this, they introduced a model that is trained on \textbf{5 million image-text pairs} of specifically marine information to better describe underwater settings this is important as it will help the \textit{LLM} gain better marine insight and language alignment this model is named \textbf{MarineGPT} \cite{20_zheng_2023_marinegpt}. This model is an essential step in explaining what detection systems find from an underwater perspective. By applying an \textit{LLM} into our system we are improving our systems effectiveness by generating summaries, this will further enhance clarity and improve efficiency by automating this step.
        
    \subsection{Summary and Gap Analysis}
    In summary, current research in underwater object detection. Improvements in \textbf{Computer Vision} techniques such as in \textbf{YOLO} and \textbf{CNN} has been apparent with advancements in detection accuracy and speed in challenging underwater environments being the solution to issues of poor visibility and small object size. Meanwhile, unsupervised \textbf{k-means clustering} methods are usually used to interpret visual data collected for further analysis. Finally, LLMs have assisted in taking over a task that was originally done by experts through enabling automatic generation of human-friendly reports based on visual data. This literature review shows examples of each of these components being applied in marine context. While much progress has been made, there is still a gap in integrating these components into one single system. This project’s main goal is doing just that, combining \textit{YOLO}-based detection, feature extraction through \textit{CNNs} and features dimensionality reduction using \textit{PCA}, \textit{K-means} clustering and visualization, \textit{LLM}-based text generation into a single, automated pipeline. The implementation of our project provides proof of concept, consistent with the observation that while integrated AI frameworks are now implementable, enhancing their performance continues to be a major focus in research\cite{30_jiang_2022_the}.

\section{Proposed Methodology}
This section explains the main methods used in building the system. To begin with, it outlines the theoretical foundations behind the chosen AI techniques. Then, it describes the system architecture, including the key modules and how they work together. Lastly, it introduces the core algorithms and explains their role in supporting underwater \textbf{data collection}, \textbf{analysis}, and \textbf{visualization}.

    \subsection{Theoretical Foundations}
    We used multiple AI techniques to support autonomous underwater exploration. Our main technique was \textbf{YOLOv12 Nano}, a lightweight \textit{Convolutional Neural Network (CNN)} model built to process video frames at once, \textbf{rapidly} and \textbf{efficiently}\cite{3_tian_2025_yolov12}. This makes it ideal for underwater exploration, where missing even a second could mean missing crucial data. After \textbf{more than 170 hours} of training on a dataset of \textbf{over 50,000 underwater frames}, the model achieved a precision score \textbf{above 0.5}, a moderate performance for a first-time detection system under challenging conditions.
    
    To analyze and group detected marine objects, we applied \textbf{K-means++ clustering} to group similar data points. This helped uncover patterns, such as clusters formed of fish detections in variety of \textbf{shapes}, \textbf{sizes}, and other \textbf{characteristics}. These clusters were based on features extracted using \textbf{ResNet50}, a deep \textit{CNN} known for its robustness in handling noisy images. Its residual connections improved training stability and helped capture finer visual details\cite{33_xiao_2024_comprehensive}.
    
    Furthermore, \textbf{Principal Component Analysis (PCA)} was applied in two steps: first, to refine the feature set for \textbf{clustering}, and second, to reduce data dimensions for clearer \textbf{2D or 3D visualizations}\cite{32_jolliffe_2016_principal}. This reduced the number of dimensions while retaining key information, helped in reducing processing overhead, a critical factor when analyzing large volumes of complex underwater imagery.
    
    Finally, we used a \textbf{Large Language Model (LLM)} through \textbf{OpenAI’s API} to generate summaries from the detection results. The LLM translated complex outputs into well-structured, readable reports, streamlining the analysis and reducing the chance of human error.
    
    By combining these methodologies, we created a practical system for the complex, ever-changing environment of the sea.

    \subsection{System Architecture}
    Our system architecture consists of multiple modules that work independently and provide their outputs on demand.
    
    \begin{enumerate}
        \item \textbf{Autonomus Vehicle}: responsible to navigate the sea to capture detections, location coordinated, and relay them to be processed through special componants, such as Computer Vision (Cameras), LiDAR, GPS Device, and more.
        \item \textbf{YOLO Object Detection Model}: \textit{You Only Look Once} is responsible for detecting objects depending on a set threshold of confedence, to be processed and analyzed, outputting a list of detected objects with bounding boxes and confidence scores.
        \item \textbf{GPS Localization Module}: responsible for coordinates processing for caching and visualizing a single or multiple detections.
        \item \textbf{Mapping Module}: responsible for visualizing either a single or multiple detections, while also providing the functionality of storing results in a centralized location.
        \item \textbf{Image Preprocessing Module}: responsible for pre-processing detections through \textit{CNN} for \textit{Feature Extraction}, providing us with a numerical representation of the detection to be analyzed (feature vectors).
        \item \textbf{Kmeans Clustering Module}: responsible for performing dimensionality reduction using \textit{PCA} to prepare the extracted features for clustering and visualization using \textit{Kmeans++}, providing us with a manageable collection of data and insights to be used to determine future detection or possible new undiscovered objects.
        \item \textbf{Large Language Model API Module}: responsible for analyzing both detections and clusters on-demand through an API, providing us with critical insights in the variety of collection of clusters and detections based on two fixed prompts.
    \end{enumerate}
    
\begin{figure}[h!]
        \centering
        \includegraphics[width=0.5\textwidth]{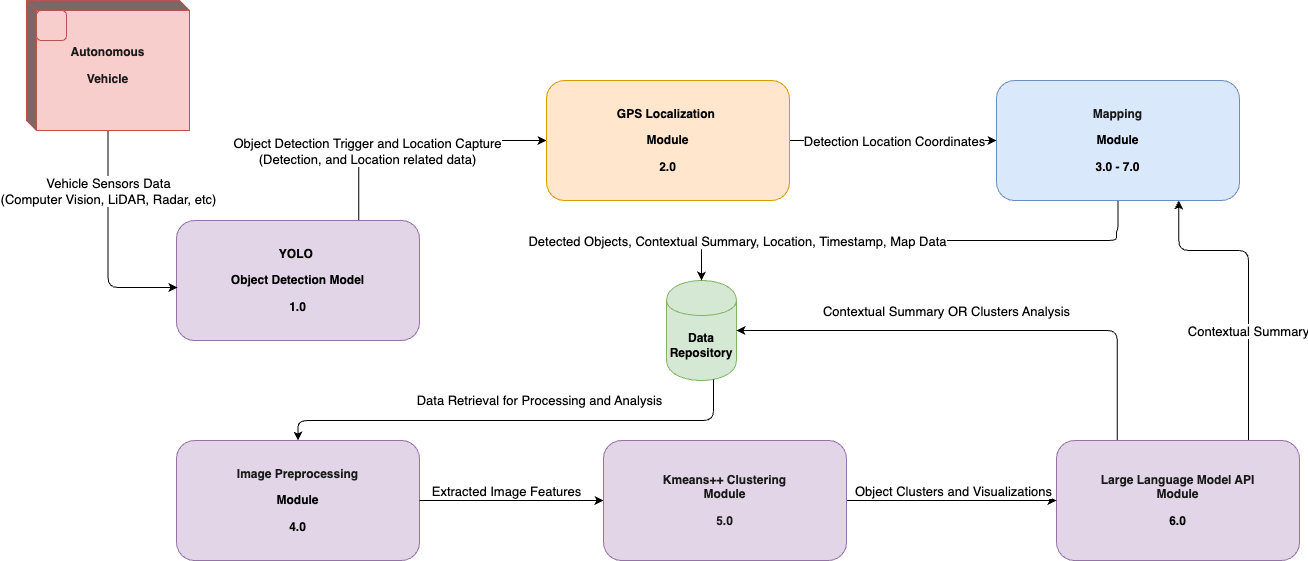}
        \caption{\textbf{System Architecture and Data Pipeline Overview:}  A detailed overview of the proposed architecture and pipeline, showing different system modules, and their data flow.} 
        \label{fig:system_pipeline}
    \end{figure}

    These modules complete each other when combined as shown in Figure~\ref{fig:system_pipeline}. Moreover, the system stores and caches relevant outputs of each module in a unified \textit{Data Repository} for reference or future use.
    
    \subsection{Algorithm Descriptions}
    Our system incorporates multiple algorithms that are detailed in Table~\ref{tab:system_algorithms}, the following is a brief description of each:
    \begin{enumerate}
        \item \textbf{YOLO}: \textit{You Only Look Once} is a state-of-the-art real-time object detection model that frames objects as a single regression problem directly predicting bounding boxes and class probabilities from full images in one evaluation. Therefore, enabling faster inference speeds, making it suitable for real-time applications\cite{10_demarcq_2023_what}.
        \item \textbf{CNN}: \textit{Convolutional Neural Networks (CNNs)} are a class of deep neural networks, mostly applied to analyze visual imagery. CNNs utilize convolutional layers, pooling layers, and fully connected layers to automatically and adaptively learn hierarchies of feauters from the input images, making them one of the most effective for tasks such as image classification and feature extraction\cite{11_lecun_2019_gradientbased}.
        \item \textbf{PCA}: \textit{Principal Component Analysis (PCA)} is widely used in dimensionality reduction technique that transform a specific set of data into orthogonal variables (principal components) that capture the most variance from the original set of data. By selecting a subset of these principal components, the dimensionality of the data can be reduced while retaining the most important information, often used in feature processing and visualization\cite{14_wold_1987_principal}.
        \item \textbf{Kmeans}: \textit{K-means} is a popular unsupervised clustering algorithm that aims to partition \textbf{n} observations into \textbf{k} clusters, in which each observation is assigned to a cluster with the nearest mean (cluster centers or cluster centroid), serving as a prototype of cluster. \textit{Kmeans++} is an initializing technique for K-means that aims to spread initial clusters centers to speed up convergence and improve final clusters quality\cite{12_arthur_2007_kmeans}.
        \item \textbf{OpenAI API}: OpenAI's API provides access to advanced \textit{Large Language Models (LLMs)} like GPT series. These models are trained on vast amounts of data, enabling them to perform a wide range of natural language processing tasks, including text generation, summerization, question answering, and understanding complex contextual information, making them effective for extracting high-level insights from structured and unstructured data, which also includes images\cite{13_brown_2020_language}.
        
    \end{enumerate}
    Combining these algorithms provides a powerful and accurate approach to achieve our scope, providing a highly flexible framework that allows for iterative refinement and the incorporation of cutting-edge algorithms.

        \begin{table}[h!]
            \centering
            \caption{Algorithms, Versions, and Libraries Implemented in the System}
            \label{tab:system_algorithms}
            \resizebox{0.5\textwidth}{!}{%
            \begin{tabular}{|l|l|l|}
                \hline
                \textbf{Algorithm} & \textbf{Version} & \textbf{Library} \\
                \hline
                YOLO & v12: Attention-Centric Object Detection & Ultralytics\\
                \hline
                CNN & ResNet50 & TensorFlow  \\
                \hline
                PCA & - & scikit-learn  \\
                \hline
                Kmeans Clustering & Kmeans++ & scikit-learn\\
                \hline
                OpenAI API  & GPT 4o Mini & openai \\
                \hline
            \end{tabular}
            }
        \end{table}

\section{Application and Implementation}
This section details the \textbf{Application and Implementation} process, focusing on the \textbf{Experimental Setup} and the \textbf{Dataset Description}. The process throughout utilized a \textit{Personal Computer} which consists of the hardware mentioned in Table~\ref{tab:system_hardware}, leveraging the \textit{CUDA} cores of an NVIDIA Graphics Card for computational acceleration and ensuring efficient utilization of other hardware components. Moreover, it was essential to prepare the image dataset structure by merging the \textbf{DeepFish} and \textbf{OzFish} datasets, ensuring no corrupted files caused compatibility issues or errors.

    \subsection{Experimental Setup}
        We fully utilized hardware components to accelerate the training of \textit{YOLO}, preprocessing of detections, performing \textit{PCA} reductions, and \textit{K-means} Clustering. For training our YOLO model, we ensured that \textbf{CUDA cores}, \textbf{32GB of RAM}, and an \textbf{NVMe Drive} were fully utilized to accelerate the training and validation process. Moreover, we leveraged the \textbf{Ryzen 5900X} and \textbf{RTX 3070 Ti} not only for training and validation but also for inferencing on video samples, providing \textbf{faster} and \textbf{more efficient} sessions that allowed us to evaluate our findings and results.
        
    \begin{table}[h!]
    \centering
    \caption{System Hardware and Software}
    \label{tab:system_hardware}
        \begin{tabular}{|l|l|}
            \hline
            \textbf{Component} & \textbf{Device Name} \\
            \hline
            Operating System & Windows 11 Pro \\
            \hline
            Central Processing Unit & AMD Ryzen 9 5900X \\
            \hline
            Graphics Processing Unit & NVIDIA GeForce RTX 3070 Ti \\
            \hline
            Random Access Memory & 32GB \\
            \hline
            Storage & 2TB NVMe - KINGSTON SNV2S2000G \\
            \hline
        \end{tabular}
    \end{table}
    
    \subsection{Dataset Description}
\begin{figure}[h]
            \centering
            \includegraphics[width=0.9\linewidth]{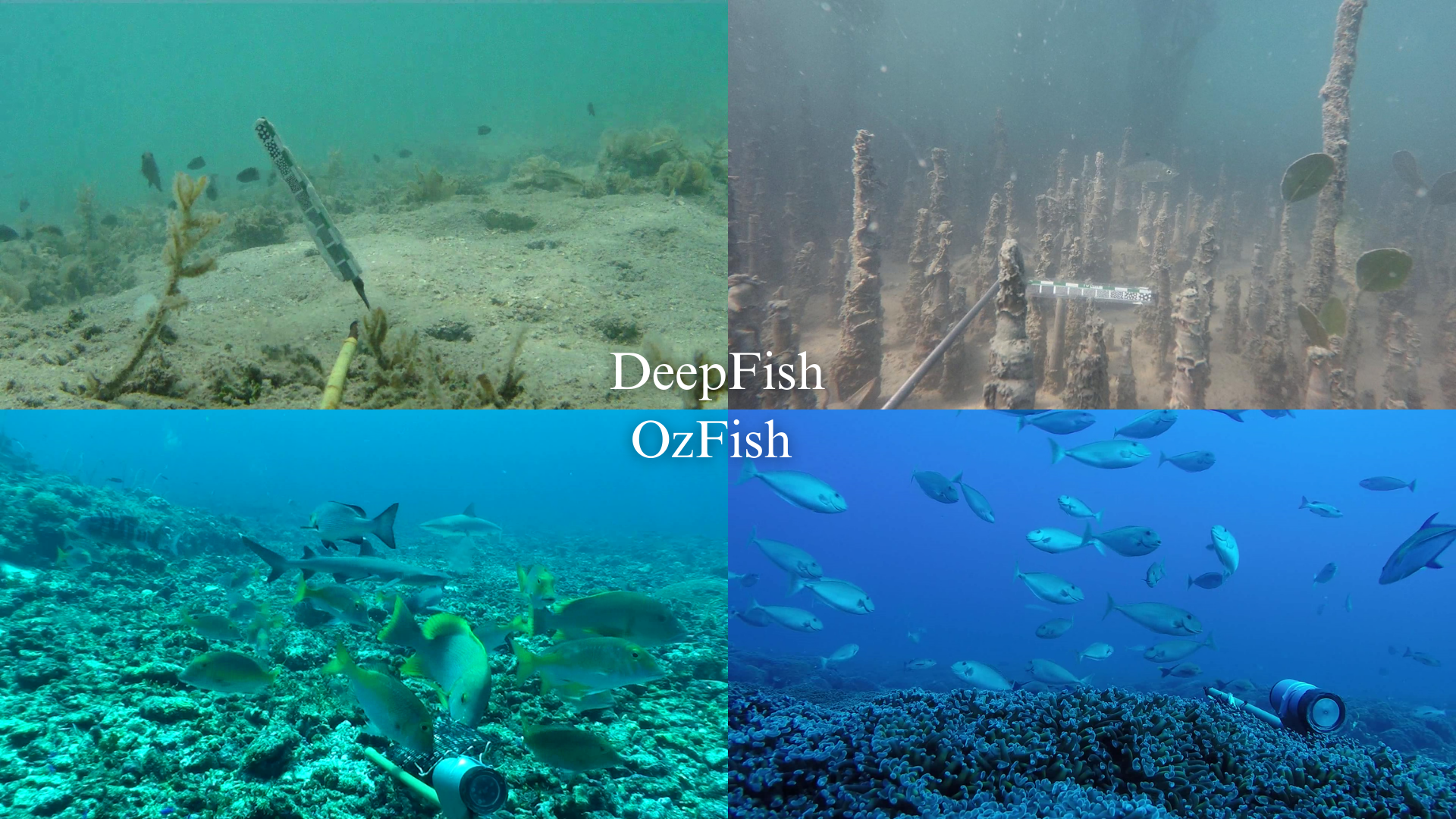}
            \caption{Image Samples From Both DeepFish and OzFish Datasets}
            \label{fig:dataset_examples}
        \end{figure}

    Our combined dataset consists of video frames derived from videos originally recorded using fixed cameras in \textbf{marine environments}. In our implementation, we \textbf{merged} the following datasets:
    \begin{enumerate}
        \item \textbf{DeepFish}: This dataset consists of frames from more than 15 habitats across the \textbf{Islands region} in \textbf{North Eastern Australia} and \textbf{Western Australia}, providing frames with bounding box annotations of fish, or no fish \cite{1_saleh_2020_a}. Furthermore, the dataset provided a structured folder that was easily \textbf{integrated} into our combined dataset without requiring further processing.
        \item \textbf{OzFish}: This dataset is part of the Australian Research Data Commons Data Discoveries program, \textbf{publicly} available for research. Similarly to \textit{DeepFish}, the videos are captured in \textbf{Australian} marine environments, providing more than 3000 video frames, with over \textbf{3000 species} and \textbf{45,000 bounding box annotations} \cite{2_australianinstituteofmarinescienceaims_2020_ozfish}. However, the dataset required some additional preprocessing steps, such as creating text files that \textbf{withhold} bounding box annotations for objects in each frame.
    \end{enumerate}
    We chose these datasets because of their ideal diversity of \textbf{environments}. \textbf{DeepFish} mostly provides visually disturbed frames resulting in low visibility, as the sample demonstrates in Fig.~\ref{fig:dataset_examples}. This is expected to enhance \textit{YOLO} performance in low-visibility situations. In contrast, \textbf{OzFish} mostly consists of clear frames which contain a high density of fish. As a result, the variety in frame quality is expected to solidify our models' performance in both high and low visibility situations.

        \subsection{Data Collection Process}
\begin{figure}[h!]
                \centering
                \includegraphics[width=0.9\linewidth]{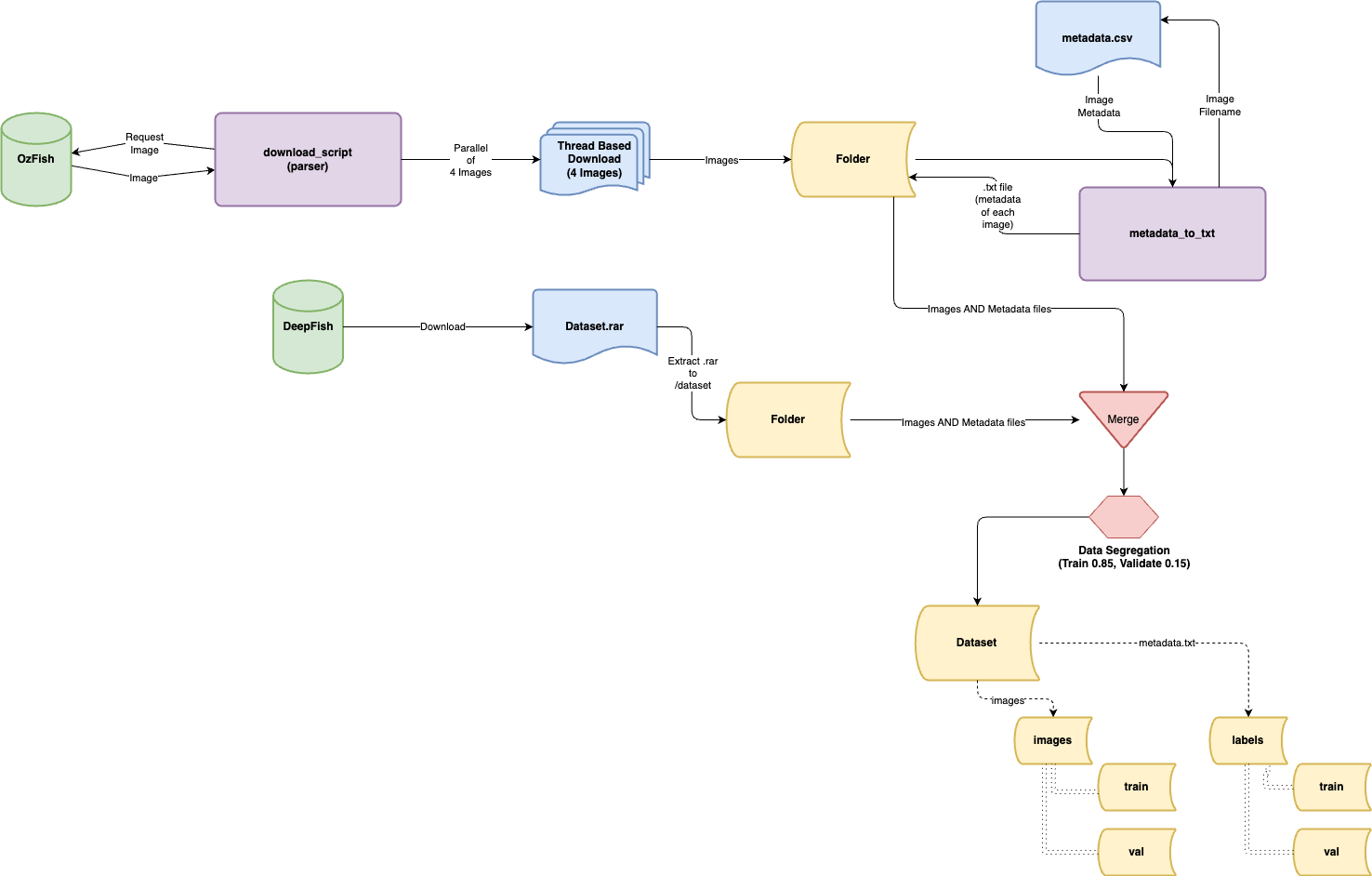}
                \caption{Dataset Building Pipeline}
                \label{fig:dataset_building}
            \end{figure}

        To achieve our final \textbf{merged} dataset, we followed best practices to ensure full \textbf{compatibility} with \textit{YOLO} in terms of the overall dataset structure and file format. Fig.~\ref{fig:dataset_building} provides a full overview of the steps that we have followed to ensure that the merge is seamlessly executed. The process included downloading the \textbf{OzFish} dataset in batches, creating corresponding text files for images downloaded through the provided \textit{metadata.csv} file, then segregating the combined \textbf{DeepFish} and \textbf{OzFish} datasets using a split of 85\% for training, and 15\% for validation, and lastly placing them in \textit{train} and \textit{val} folders.

        \subsubsection{Metadata Analysis}
        The merged dataset consisted of more than \textbf{496 species} of fish from multiple habitats around \textbf{Australia}. Additionally, the dataset contained fish of multiple sizes, colors, and shapes. This variety results in a diverse set of frames that the \textit{YOLO} model might benefit from for improved generalization in detections. However, this may also result in an imbalance in the samples that \textit{YOLO} would observe\cite{28_crasto_2024_class}. Furthermore, these dataset frames are in a \textbf{resolution of 1920x1080} which provides a significant amount of detail for the annotated instances, such as object color, shape, precise size, and texture as shown in Table~\ref{tab:dataset}.
        
        \subsubsection{Data Visualization}
\begin{figure}[h]
            \centering
            \includegraphics[width=0.9\linewidth]{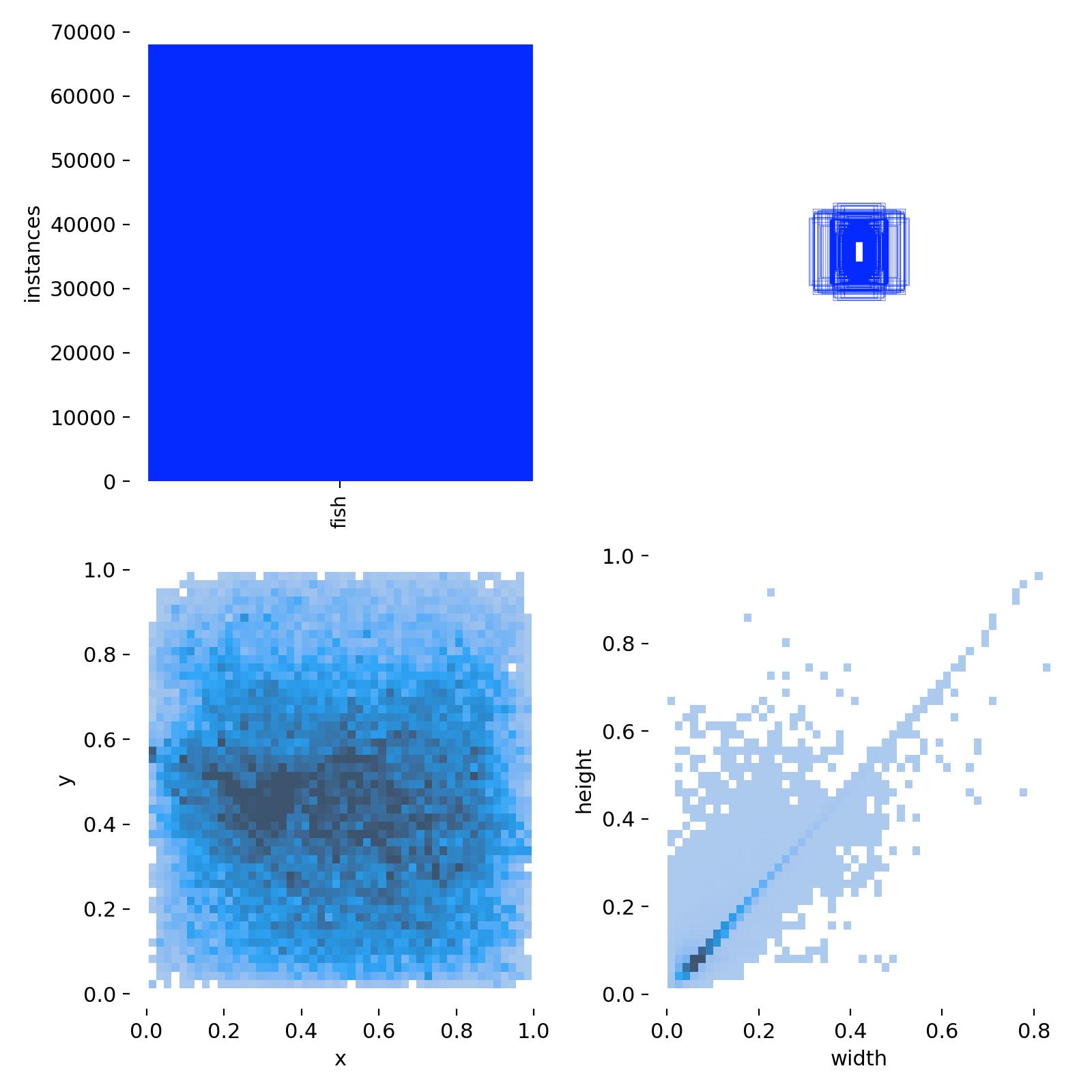}
            \caption{Distribution of Dataset Instances and Bounding Box Properties}
            \label{fig:labels_data_vis}
        \end{figure}

        Our data consists of manually annotated bounding boxes in a variety of sizes for one or multiple instances in
        an individual frame. Our merged dataset consisted of more than \textbf{60k instances} across both datasets \textbf{DeepFish} and \textbf{OzFish}. Moreover, these instances are in bounding boxes in location within a frame of \textbf{x} and \textbf{y} as shown in Fig.\ref{fig:labels_data_vis}, they also vary in their width and height which corresponds to the annotated instance size.

        \begin{table}[h]
            \centering
            \caption{Dataset Composition (Image Count, Resolution, and Format)}
            \label{tab:dataset}
            \begin{tabular}{|l|l|l|l|}
                \hline
                \textbf{Dataset} & \textbf{Image Count} &\textbf{Resolution} &\textbf{Format} \\
                \hline
                DeepFish & 4,505 & 1920 x 1080 & JPEG\\
                \hline
                OzFish & 51,217 & 1920 x 1080 & PNG \\
                \hline
                \textbf{Total} & 55,722 & - & -\\
                \hline
            \end{tabular}
        \end{table}
        
\section{Experiments and Results Analysis}
This section will discuss the results of our experiment and what they reveal. Starting from evaluation metrics, an outline of key metrics used to measure performance. Then we will discuss our results in detail. Finally, we will interpret the result.

    \subsection{Evaluation Metrics}
    In this section, we evaluate the performance of the system using metrics to measure \textit{YOLO’s} detection accuracy, \textit{PCA’s} dimensionality reduction, and \textit{K-means} clustering effectiveness. These metrics deliver insights into how well the system performed across different stages of analysis. A description of each metric is listed below:
\begin{itemize}
    \item \textbf{mAP@0.5:} Mean Average Precision \textbf{at 0.5 IoU}, which calculates the average precision for each class when the predicted boxes match the real boxes by \textbf{at least 50\%}, displaying the overall detection accuracy.
    \item \textbf{Precision:} Proportion of \textbf{correct detections} out of all \textbf{positive predictions}, showing the ability for the model to avoid false positives.
    \item \textbf{Recall:} Proportion of \textbf{actual positives correctly predicted} reflecting the model’s ability to find all objects in an image, which lessens the chances of missed detections.
    \item \textbf{Components:} The amount of \textbf{reduced dimensions} in PCA that store important information while reducing dimensionality to assist with \textbf{clustering} and \textbf{visualization}.
    \item \textbf{Kmeans:} The \textbf{number of clusters} chosen to group by similarity, revealing underlying trends in the data such as variety of objects \textbf{sizes}, \textbf{shapes}, and other \textbf{characteristics}.
\end{itemize}

        \begin{table}[htbp]
            \centering
            \caption{Performance Metrics and Parameters Used}
            \label{tab:metrics}
            \resizebox{0.5\textwidth}{!}{%
            \begin{tabular}{|l|l|l|}
                \hline
                \textbf{Metric} & \textbf{Algorithm} & \textbf{Description} \\
                \hline
                mAP & YOLO & Measures Mean Average Precision between predicted and actual labels \\
                \hline
                Precision & YOLO & Proportion of correct detections \\
                \hline
                Recall & YOLO & Proportion of actual positives correctly predicted \\
                \hline
                Componants & PCA & Number of Dimensions Reduction \\
                \hline
                K & Kmeans++ & Optimal Number of Clusters \\
                \hline                
            \end{tabular}%
            }
        \end{table}
    
    \subsection{Experimental Results}
    Our fish detection system was evaluated using multiple experimental tests. The results from each test revealed our model's capabilities and limitations, which provided us with insight into its performance in underwater image detection. The YOLO model achieved an \textbf{mAP@0.5} score of \textbf{0.512}, indicating its overall detection performance. The precision score was \textbf{0.535}, while the recall score reached \textbf{0.438}.
    In addition to \textit{YOLO}, we applied \textit{PCA} and \textit{K-means} to further process the detection results. \textit{PCA} was used for dimensionality reduction, and we have chosen to keep 0.98 cumulative explained variance, which resulted in \textit{199 components}. The \textit{K-means} clustering method was applied, and the number of clusters varied depending on the total number of detections found during analysis. This allowed the system to adapt to different sets of detection results.

            \begin{table}[htbp]
            \centering
            \caption{Summary of Key Results and Parameters}
            \label{tab:Results}
            \resizebox{0.5\textwidth}{!}{%
            \begin{tabular}{|l|l|l|}
                \hline
                \textbf{Algorithm} & \textbf{Metric} & \textbf{Result} \\
                \hline
                YOLO & mAP@0.5 & 0.512 \\
                \hline
                YOLO & Precision & 0.535 \\
                \hline
                YOLO & Recall & 0.438 \\
                \hline
                PCA & components & 900 OR 0.98 Cumulative Explained Variance \\ 
                \hline
                Kmeans & num of clusters & depending on input sum of detections \\
                \hline
            \end{tabular}%
            }
        \end{table}
        
    \subsection{Discussion of Results}
    In this section, we will evaluate our algorithms' performance and interpret their results. First, we will examine \textit{YOLO's} performance in terms of detection accuracy and confidence thresholds that affect the extraction of image crops. We then analyze results from the \textit{CNN} used for feature extraction and \textit{PCA} for dimensionality reduction. Next, we discuss the clustering results from \textit{K-Means}, explaining what each cluster represents and what attributes characterize different clusters. Finally, we will interpret the results of our \textit{LLM} API integration and its role within our system. Together, these analyses illustrate both the strengths and limitations of our approach and suggest avenues for future improvement.
    
        \subsubsection{\textbf{YOLO Object Detection Performance}}
        We begin by evaluating the performance of \textit{YOLO} on the fish detection task. Table~\ref{tab:Results} summarizes the results obtained during the training stage, such as Mean Average Precision, Precision, and Recall. Our model achieved a \textbf{mAP@0.5 of 0.512}, \textbf{Precision of 0.535}, and \textbf{Recall of 0.437} at \textbf{epoch 300}, meaning it is able to detect and distinguish \textbf{more than 50\%} of the annotated objects from the dataset it was trained on. Moreover, the model showed promising inference speed ranging \textbf{from 2.0ms to 5.5ms}, which is especially beneficial in marine environments. Additionally, to verify our model's performance, we collected four videos captured by individuals and published on \textbf{YouTube}. These videos were \textbf{hand-picked} specifically to contain fish \textit{species} present in our dataset\cite{4_filmingaustralia_2022_sea, 5_naturerelaxationfilms_2020_new}. 
        
        Our results showed that the model was able to detect multiple fish species with various confidence scores ranging from 0.85 to 0.1, as shown in Fig.\ref{fig:inference_results}. These results are promising, especially considering we used newly picked frames for inferencing, with clear image crops to be processed. However, multiple attributes affected the detection of fish. These attributes were the \textbf{lighting} and the \textbf{angle} of the fish itself within the frame; such attributes required us to lower the confidence threshold to maintain detection. Furthermore, the metrics observed in Table\ref{tab:Results} might not represent the ideal performance of a detection model, but the class imbalance within the dataset negatively affected the training process, as speculated in the \textit{Metadata Analysis} section. This raises the need to mitigate this imbalance with advanced techniques such as \textbf{Sampling} and \textbf{Augmentation}\cite{28_crasto_2024_class}.
            
        \subsubsection{\textbf{Insights from CNN and PCA}}
        Following the object detection stage, we move on to preprocessing and preparing our detection crops for analysis. These crops are first used as input for a \textit{CNN} model for feature extraction, which results in a numeric vector representing the characteristics of that detection. This vector is then flattened. The result of this stage is a set of preprocessed numeric representations of the detection crops that are ready to be clustered and visualized.
        
        Throughout the execution of this stage, we observed that \textit{PCA} component selection varies depending on the number of detection crops used as input. For instance, when we attempted to reduce a vector of \textbf{more than 2,000 crops} while preserving 98\% variance, we encountered multiple computing errors, likely because of our hardware constraints. As indicated in Table~\ref{tab:Results}, the chosen number of components might be 900, or alternatively, selected to preserve 98\% (0.98) of the variance, with the specific value determined by the input size. When specifying 98\% variance preservation, the returned components range from \textbf{150 to 800}, resulting in a much more manageable number of components. This process results in a set of numerical data that preserves the information of individual crop characteristics, and also helps mitigate errors due to our hardware constraints\cite{14_wold_1987_principal}.
        
        Throughout the execution of this stage, we observed that \textit{PCA} component selection varies depending on the number of detection crops used as input. For instance, when we attempted to reduce a vector of \textbf{more than 2,000 crops} while preserving 98\% variance, we encountered multiple computing errors, likely because of our hardware constraints. As indicated in Table~\ref{tab:Results}, the chosen number of components might be 900, or alternatively, selected to preserve 98\% (0.98) of the variance, with the specific value determined by the input size. When specifying 98\% variance preservation, the returned components range from \textbf{150 to 800}, resulting in a much more manageable number of components. This process results in a set of numerical data that preserves the information of individual crop characteristics, and also helps mitigate errors due to our hardware constraints\cite{14_wold_1987_principal}.
        
       \subsubsection{\textbf{Insights from K-Means Analysis}}
\begin{figure}[h]
            \centering
            \includegraphics[width=1\linewidth]{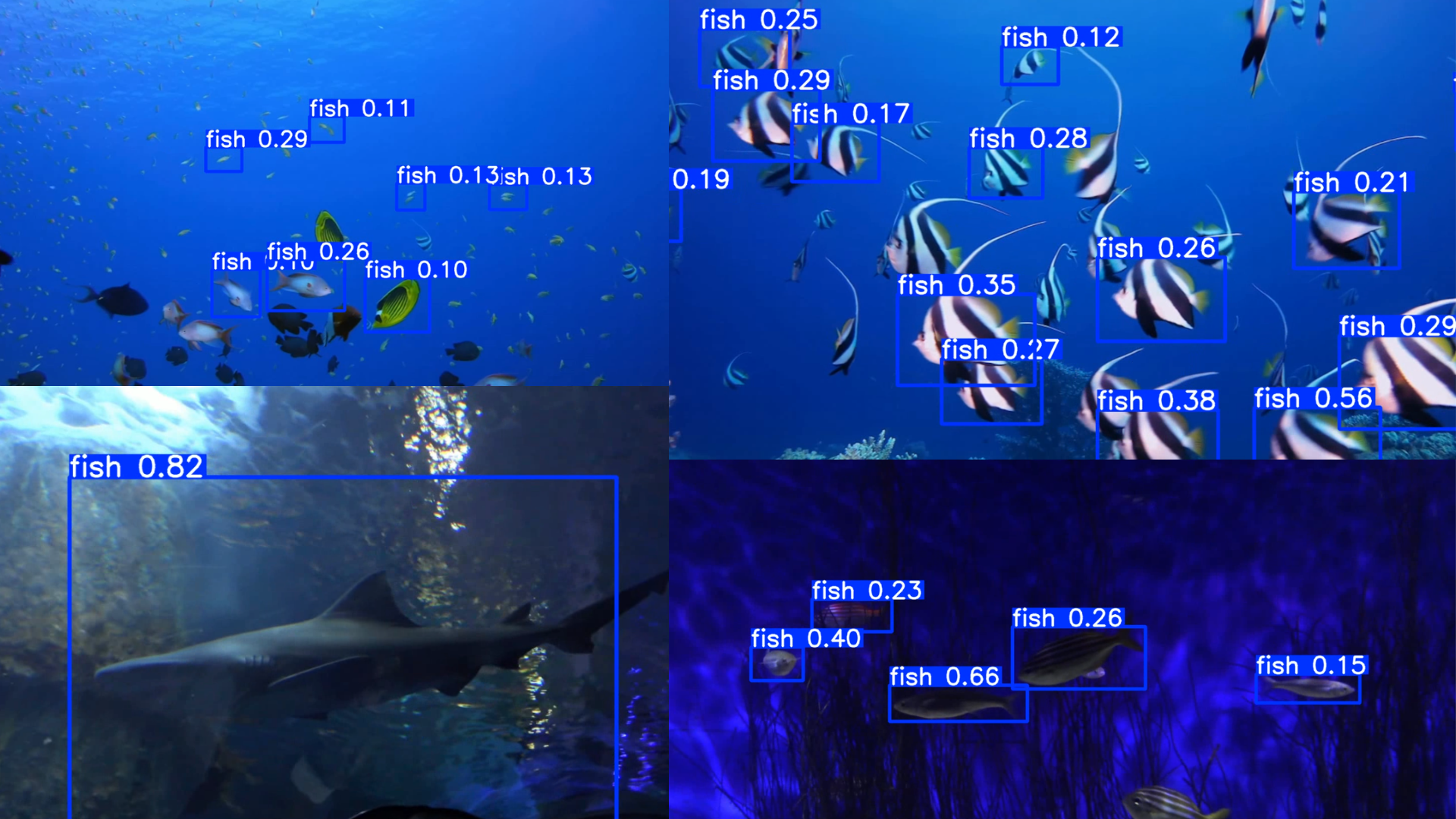}
            \caption{Examples of Object Detection Inference Results}
            \label{fig:inference_results}
        \end{figure}

        Following the \textit{Feature Extraction} and \textit{PCA} Reduction stage, we continue with both clustering and visualizing our results for \textbf{interpretation}. We've used \textit{K-Means} for clustering and \textit{Plotly} for visualizing in both 2D and 3D, leveraging its features that include displaying metadata of individual data points of a cluster, on hover. This combination enabled us to uncover and interpret many patterns within our detections set gathered from our \textit{inference} shown in Fig.\ref{fig:inference_results}.
        
        During our implementation, we observed that the \textbf{number of k of clusters} to use is highly dependent on the number of crops used as an input, and their different characteristics, such as lighting, object angle, object movement, and object shape, as indicated in Table~\ref{tab:Results}. During our execution, the optimal number of clusters was \textbf{27 for 687 detection crops}. To further examine them, we \textbf{hand-picked 4 different crops} to be processed and assigned to clusters. Moreover, we've ensured that these picked crops offer \textbf{different characteristics} of lighting, object angle, and species. After processing the selected crops and assigning them to clusters, we started to examine these assigned clusters individually and comparing them to our testing crops as shown in Fig.~\ref{fig:labels_data_vis}. From this examination, we've observed that the predicted clusters were formed, and our crop picks were correctly assigned. Most of the other crops present were \textbf{relatable}, \textbf{similar}, or \textbf{identical} in their characteristics. However, there was a minor observation that was interesting: in the prediction of Cluster 1, the tested crop was not identical in terms of fish species, but it contained very similar  frame lighting, object shape, and object angle to the other crops. This might present a challenge in some cases, particularly if the goal is to have fully pure clusters of a specific species. In contrast, we can leverage the observations found in cluster 1 to our advantage in discovering and assigning new unknown species or objects to clusters that share similar characteristics to those typical of a certain detection crop.

            \begin{figure}[h]
                \centering
                \includegraphics[width=0.9\linewidth]{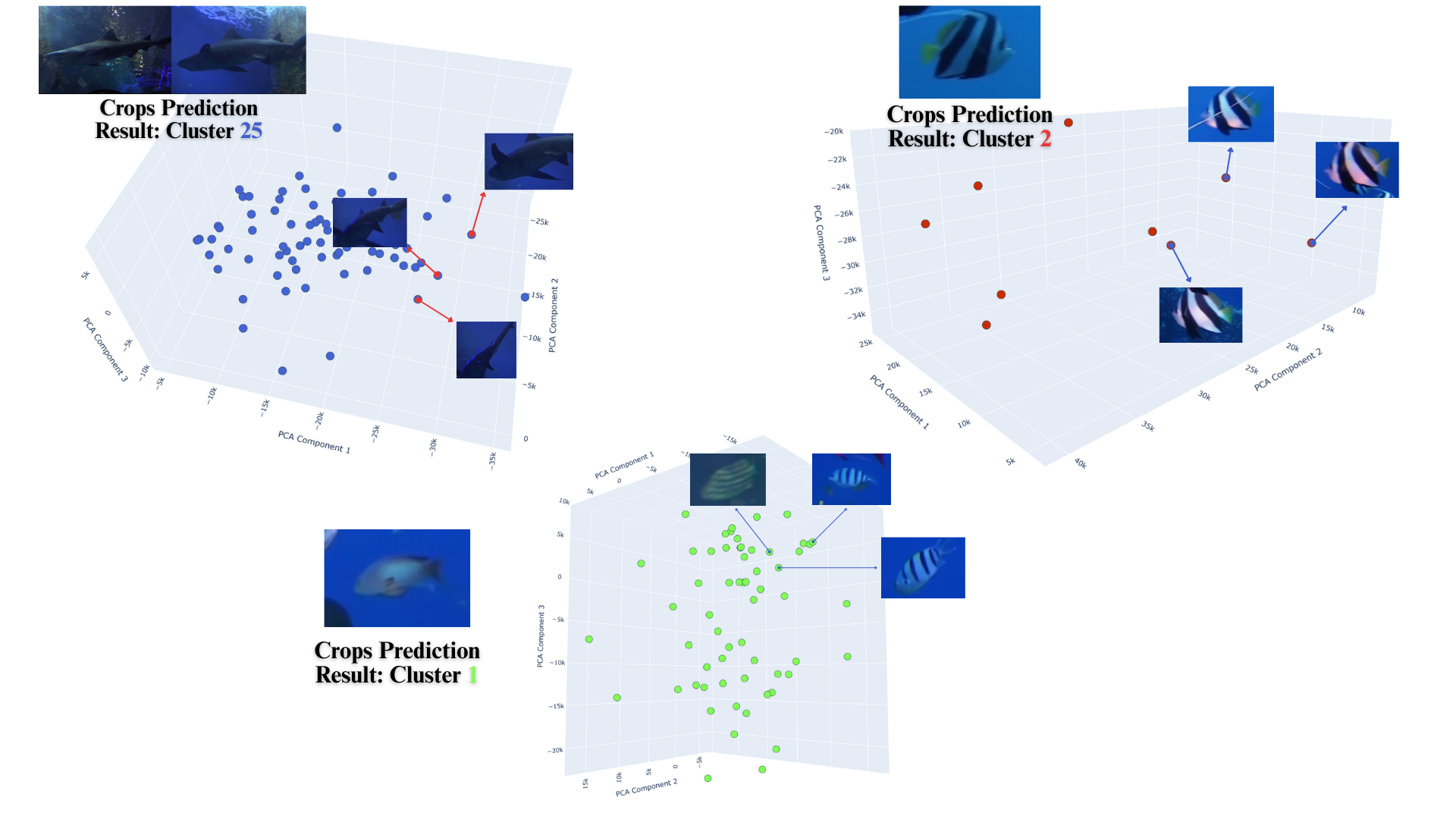}
                \caption{Visualization of Test Crop Assignments within PCA-Reduced Feature Space Clusters}
                \label{fig:prediction_results_and_clusters_visual}
            \end{figure}
        
        \subsubsection{\textbf{Contributions from the LLM API and Location Coordinates}}
        Lastly, we discuss the integration and role of the \textit{LLM API} of \textbf{GPT-4o-Mini} in our system, mentioned in Table~\ref{tab:system_algorithms}. This integration provides the unique ability to \textbf{analyze} and \textbf{summarize} both detections and clusters through requests that contain a \textbf{special prompt} tailored to our specific context of fish detection. Moreover, each request may also contain relevant attachments, such as an image crop for analysis or an html file as a string format that visualizes our clusters results. Combining the special prompt and the relevant attachments provided the \textit{LLM} with the necessary context needed to complete the tasks of \textbf{Detection} or \textbf{Cluster Analysis}.
        
        In our implementation, we've tested multiple prompts to ensure that the response is relevant to our context of \textbf{Sea Exploration}. The combination of prompts tested included mentioning to the \textit{LLM} that the attached crop or cluster is from \textit{YOLO} Detections in sea environments, instructions on describing the detected \textbf{object's shape}, \textbf{size}, \textbf{texture}, any \textbf{discernible patterns}, and \textbf{places} where the object might exist, live, or is commonly seen \cite{31_berasi2025textexploringcompositionalityvisual}. Moreover, we combined the \textit{LLM} with the visualization of \textbf{coordinates} retrieved from individual detections. These coordinates may also be used as input for the \textit{LLM} to further support its \textbf{analysis} and \textbf{summarization}. Moreover, these analyses and summaries may be visualized, and included within \textbf{individual data points} in a map visual, or cached for future reference or study.
        
\begin{figure}[h]
            \centering
            \includegraphics[width=1\linewidth]{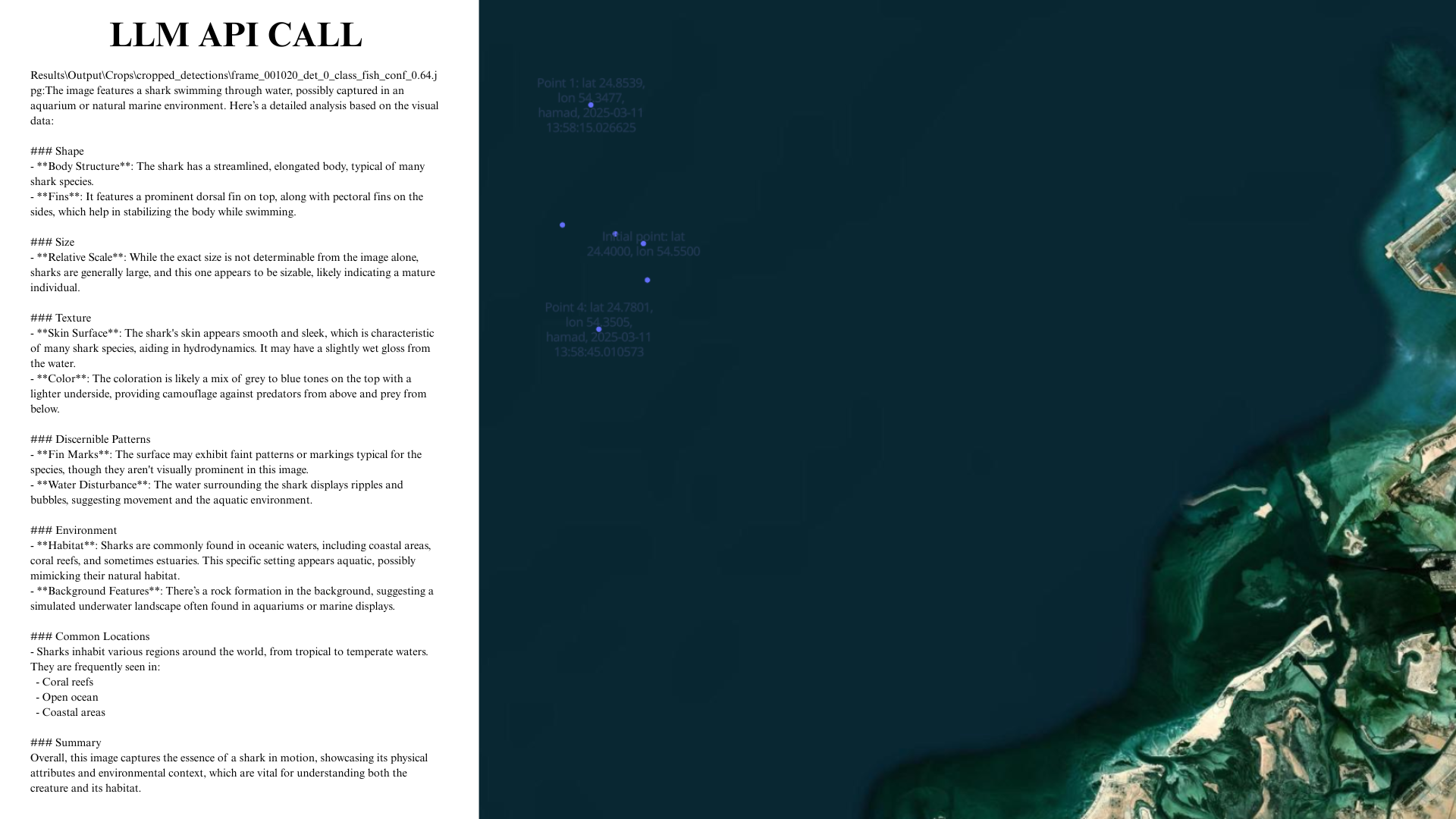}
            \caption{Example LLM Analysis of a Detection and Map Visualization of Detection Locations}
            \label{fig:llm_api_and_location_visual}
        \end{figure}

        Fig.~\ref{fig:llm_api_and_location_visual} provides a sample of results obtained during our implementation. We used our \textbf{special prompt} and attached a detection of a Shark to the request. The \textbf{LLM API} response was as expected; it included all of the relevant information for that Shark detection. This included a brief summary, description of shape, size, texture, discernible patterns, environment, and the common locations where a shark might exist. These results are promising for \textbf{real-world deployment}. However, we faced a couple of challenges in attaching the html file related to our clusters visualization to the \textit{LLM API}. These challenges included the \textbf{long structure of the file}, resulting in a \textbf{token-intensive request} that is beyond our resources. Future enhancements to this implementation may include running a \textbf{local mini model of an LLM} specialized in \textbf{marine biology knowledge} to support this feature within a system. Lastly, in Fig.~\ref{fig:llm_api_and_location_visual}, \textit{Plotly} in combination with \textit{Dash} provided the ability to visualize a simulation of a detection and multiple detections in the \textbf{Arabian Gulf}. This simulation provides insights into how a real-world deployment can be visualized and interpreted on a larger scale.

    Overall, our implementation demonstrated a working concept of a system that could be further enhanced for real-world deployment. This involved achieving relevant detections using \textit{YOLO}, employing a \textit{CNN} for feature extraction from crops, reducing feature dimensionality through \textit{PCA} reduction while preserving characteristics of detected objects, and gaining insights from an \textit{LLM API} regarding detections and clusters, with the incorporation of coordinate collection and visualization. However, there is room for improvement in \textit{YOLO's} performance, further enhancing the processes of feature extraction and \textit{PCA} reduction to preserve information more efficiently. Additionally, adjusting the hyperparameters of \textit{K-Means} could help achieve optimal and ideal clusters for interpretation; these adjustments might be aimed for either pure or similar-in-characteristics clusters. Moreover, finding a more advanced integration for the \textit{LLM} model to handle more tasks than just detection and cluster analysis is a potential future step.
    
\section{Conclusion}
The project successfully developed and implemented an underwater AI system which detects, clusters, and interprets marine life data. The system used \textbf{YOLOv12 Nano} to detect and distinguish \textbf{over 50\% of annotated objects} under challenging marine conditions, with a \textbf{mAP@0.5 of 0.512}, a \textbf{precision of 0.535}, and a \textbf{recall of 0.437} at \textbf{epoch 300}. The system maintained consistent inference speeds between \textbf{2.0ms and 5.5ms} and processed external video data to validate its performance. The \textit{CNN} model extracted numeric features, which \textit{PCA} processed to reduce \textbf{over 2,000 crops while maintaining 98\% cumulative explained variance}, before selecting 900 components for efficient dimensionality reduction. The \textit{K-means} clustering algorithm grouped \textbf{687 detection crops} into \textbf{27 clusters}, while detection patterns depended on lighting conditions, object size, and shape. The system combined an \textit{LLM} (GPT-4o Mini) to generate structured reports by processing detection outputs and location data. The LLM generated effective detection summaries through specific prompts and delivered comprehensive detections location and characteristic analysis, which supports both real-world deployment and enhanced data interpretation capabilities.

    \subsection{Challenges and Limitations}
    While the integration of YOLOv12, K-means clustering, and an LLM for summarization has proven to be implementable to an AI-powered underwater detection system, several challenges that affected certain aspects of it’s implementation were identified. Underwater environment such as \textbf{opaqueness} and \textbf{poor lighting} has greatly affect the accuracy of detection which in return causes missed detections or inaccurate readings. A challenge that most data scientists struggle with is the \textbf{curse of dimensionality} which refers to how models perform poorly when handling datasets with too many features was a critical challenge in our case; the large number of extracted features made it hard to keep the most informative ones leading to difficulty in maintaining reliability. Likewise, the model struggled with highlighting features clearly, leading to wrong or missed detections. Finally, we struggled with YOLO model selection as we needed to decide which YOLO version best fits underwater detection.
    
    On the other hand, several limitations that affected certain aspects of its implementation were identified. \textbf{Hardware limitations} were our primary issue due to the lack of availability of computational resources; this lead to accuracy limitations. Dataset limitations due to the limited availability of high-quality annotated underwater datasets which made data preparation very difficult. \textbf{Class Imbalance} was another reason our model struggled with generalization and accuracy of detection as certain classes were underrepresented. \textbf{Image quality limitations}: due to low lighting, blur, color distortion, or unclear images within the dataset impacted training effectiveness and reduced detection performance. Time constraints due to the need to perform \textit{YOLOv12} detection processes, \textit{k-means} clustering, and \textit{LLM} summarization under a limited amount of time.
    
    \subsection{Future Work}
    Although our system's implementation has demonstrated a working concept of AI integration for autonomous underwater object detection, clustering, and reporting, it also uncovered a set of directions that future implementations might focus on for further advancements and refinements. The challenging nature of underwater environments and the complexities of AI integration present new research opportunities, informing these directions. This section outlines key directions for future work aimed at enhancing the overall system's performance.

        \begin{enumerate}
        
            \item \textbf{Improving Detection of Challenging Cases:} There is a need to address class imbalance in the dataset, which affects the training process and performance of YOLO. Future work may explore different combinations of datasets or advanced methods such as \textbf{sampling} and \textbf{augmentation} \cite{28_crasto_2024_class} to achieve better performance metrics for YOLO.

            \item \textbf{Advanced Methods to Handle Curse of Dimensionality:} Handling the curse of dimensionality remains one of the major challenges in implementing systems that depend on multi-dimensional features extracted through CNNs. Future work may include exploring different custom methods specifically tailored for the unique characteristics of features extracted from sea-related organisms. 
                        
            \item \textbf{Expanding LLM Integration:} Future work will explore advanced methodologies for LLM integration to operations beyond simple analysis of detections or clusters. These advanced integrations could include generating real-time reports on detections or providing support for analyzing detection to cluster assignments.

            \item \textbf{Expanding Application on Other Object Classes:} Expanding the set of classes for the model to detect objects beyond fish would provide critical insight into the overall impact of such system. These other classes may include sea-related organisms such as seahorses, shrimps, crabs, snakes, or even plants. Expanding this application would further support the broader goals of exploration and diversify the system's overall purpose.
            
            \item \textbf{Considering Real-world Deployment:} Future work will involve exploring advanced methodologies related to model quantization and deployment on an autonomous vehicle. This direction aims to provide in-depth insights into how such systems perform in real-world environments and identify potential challenges that might arise.
            
        \end{enumerate}

    By following the directions outlined above, significant advancements will be achieved in the field of autonomous underwater exploration. Addressing these challenges is crucial for developing a more capable, reliable, and versatile AI-powered system that can unlock the vast potential for scientific discovery and environmental monitoring in the world's oceans.

\section*{Acknowledgements}
We would like to express our sincere gratitude to Doctor \textbf{Mohammad Tubishat}, and Doctor \textbf{Edmund Evangelista} for their invaluable guidance, support, and insightful discussions throughout this work. Their expertise and encouragement were instrumental in the completion of this project.

We also thank \textbf{Australian Institute of Marine Science (OzFish)} and \textbf{A. Saleh, I. H. Laradji, D. A. Konovalov, M. D. Bradley, D. Vazquez, and M. Sheaves (DeepFish)} for providing access to their datasets, which were essential for this research.

\bibliographystyle{IEEEtran} 

\bibliography{References} 

\end{document}